\pdfoutput=1

\documentclass[11pt]{article}

\usepackage[]{acl}

\usepackage{times}
\usepackage{latexsym}

\usepackage[T1]{fontenc}

\usepackage[utf8]{inputenc}

\usepackage{microtype}

\usepackage{inconsolata}

\usepackage{graphicx}
\usepackage{enumerate}
\usepackage{subcaption}
\title{Learning to Route for Dynamic Adapter Composition in Continual Learning with Language Models}

\author{%
Vladimir Araujo, Marie-Francine Moens, Tinne Tuytelaars\\
KU Leuven\\
\texttt{vladimir.araujo@kuleuven.be}
}

\begin{document}
\maketitle

\begin{abstract}
Parameter-efficient fine-tuning (PEFT) methods are increasingly used with pre-trained language models (PLMs) for continual learning (CL). These methods typically involve training a PEFT module for each new task and employing similarity-based selection to route modules during inference. However, they face two major limitations: 1) interference during module training with already learned modules and 2) suboptimal routing when composing modules. In this paper, we present L2R, a method that isolates the training of new PEFT modules to ensure their task specialization. L2R then learns to compose the learned modules by training a network of routers that leverages a small memory containing examples of previously seen tasks. We evaluate our method in two CL setups using various benchmarks. Our results demonstrate that L2R provides an effective composition of PEFT modules, leading to improved generalization and performance compared to other methods.
\end{abstract}

\section{Introduction}

CL aims to continuously learn new tasks without forgetting previously learned knowledge from old tasks \cite{MCCLOSKEY1989109}. Recent advancements in PLMs and PEFT methods have shown potential in CL for NLP \cite{zhou2024continual}.

PEFT methods, like prompts or adapters, are lightweight modules that can be trained for downstream tasks while keeping the PLM frozen. Recent work shows that these methods can be competitive with, or even superior to, full fine-tuning of PLMs \cite{li-liang-2021-prefix,hu2022lora}. Additionally, research \cite{wang-etal-2023-rehearsal,wang2024rehearsalfree,Wang_2022_CVPR,Smith_2023_CVPR} demonstrates their effectiveness in CL.

However, PEFT approaches for CL have two significant limitations: 
1) When a new task arises, a new PEFT module is added, and the previously learned ones remain frozen but activated \cite{razdaibiedina2023progressive}. Previous modules may interfere with optimal task knowledge acquisition by the current PEFT module.
2) They rely on a similarity-based selection mechanism \cite{wang2024rehearsalfree} that may not accurately represent a routing function that effectively composes the PEFT modules.

To address these issues, we introduce \textbf{L}earning \textbf{t}o \textbf{Ro}ute for dynamic PEFT composition (L2R). L2R is a simple but effective method that trains PEFT modules in isolation to avoid interference from previous knowledge when learning new tasks \cite{wang-etal-2023-rehearsal}. Then, it uses previous examples stored in a memory to learn a routing function for dynamically composing PEFT modules during inference, a process analogous to local adaptation \cite{NEURIPS2019_f8d2e80c}.
Our extensive evaluations show that L2R is competitive against other PEFT-based methods across benchmarks and CL setups.

Unlike existing methods that focus on developing routing mechanisms within the training phase, our method learns to route adapters prior to test-time inference.
From a practitioner's perspective, the ultimate goal of CL is to develop a model that can learn new tasks while maintaining high performance \cite{10.1007/978-3-030-58536-5_31}.
Our method addresses this by incrementally adding PEFT modules for new tasks.
Additionally, in real-world applications, systems are primarily constrained by computational and time budgets rather than storage \cite{10204648,verwimp2024continual}.
By leveraging a memory along PEFT, our approach achieves a better trade-off between performance and efficiency.

\begin{figure*}[t]
\centering
\includegraphics[width=0.95\linewidth]{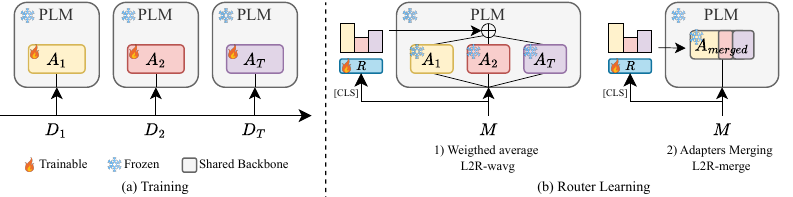} \hfill
\caption{Overview of the L2R method. (a) Adapters $A$ attached to the backbone are sequentially trained on a series of tasks $D$. Each adapter undergoes isolated training to prevent interference. (b) Before performing inference, our method utilizes a memory $M$ to learn a routing function $R$, facilitating composition either by 1) computing a weighted average of the adapters' outputs or 2) merging the parameters of the adapters.}
\label{figure:model}
\end{figure*}

\section{Related Work}

\paragraph{Continual Learning with PLMs}

CL in NLP has grown with the advent of robust PLMs from pre-training, beneficial for CL \cite{zhou2024continual}. Existing NLP approaches include replay-based \citep{NEURIPS2019_f8d2e80c,Araujo_2022_CVPR}, meta-learning-based \citep{wang-etal-2020-efficient}, amount others.

Recent techniques use PEFT for CL, adding new modules as tasks arise while freezing previous ones to maintain a robust backbone and leverage past knowledge \cite{razdaibiedina2023progressive}. This approach has been extended to learn the composition of PEFT modules based on similarity \cite{wang2024rehearsalfree}. While effective, these methods may suffer from interference during learning new modules and suboptimal module composition during inference.

Our work focuses on the PEFT setup. We rely on the parameter isolation strategy \cite{wang-etal-2023-rehearsal,cheng2024dynamic} to let the PLM learn task-specific modules independently. 
We also adopt a memory as in replay-based methods, not to rehearse examples but to learn a routing function for effective module composition before inference.

\paragraph{Local Adaptation}

Memory-based parameter adaptation (MbPA) \cite{sprechmann2018memorybased}, also known as local adaptation, allows a model to deal with changes and shifts in data distributions. In CL, it leverages labeled data stored in memory for brief fine-tuning before making predictions to improve model performance \cite{NEURIPS2019_f8d2e80c}.

Our work builds on the concept of local adaptation to maintain model generalization and performance across tasks in CL. Rather than retraining model components, we introduce a new component (a router) that learns to dynamically compose appropriate PEFT modules based on input.

\paragraph{Learning to Route}

In mixture-of-experts (MoE) models, routing activates subnetworks based on their specialized capabilities for prediction \cite{shazeer2017}. Recently, adapters—a class of PEFT methods—have been employed to implement MoE-style models \cite{wang-etal-2022-adamix} and routing functions \cite{ponti-etal-2023-combining}.

Existing approaches in CL use task centroids \cite{cheng2024dynamic}, learnable task vectors \cite{wang2024rehearsalfree}, or record task distributions \cite{wang-etal-2023-rehearsal} to select modules. However, these act as a proxy for routing and may not effectively direct input to modules based on their specialties.

Our work builds on studies showing the benefits of a well-learned router in MoE models \cite{dikkala-etal-2023-benefits}. We propose a router learning approach that takes place prior to inference and enables dynamic utilization of modules by the model.

\section{Method}
\label{sec:method}

We consider a CL setup for NLP \cite{NEURIPS2019_f8d2e80c}, where a model needs to learn from a sequence of $T$ tasks: $ D = \{D_1, ..., D_T \} $. Each task $t$ consists of a new data distribution $D_t = (X_t,Y_t)$, where $X_t$ are the input instances and $Y_t$ are the labels.
A model in this setup consists of a PLM that processes input $x$ and a classifier for predicting class $y$.
We extend this setup with PEFT modules, specifically adapters, which consistently achieve superior predictive performance than other techniques in CL \cite{wistuba2024choice}.

\paragraph{Task-specific Adapters}
L2R employs a set of adapters $ A = \{A_1, ..., A_T \} $ that learn task-specific knowledge from a data stream (\autoref{figure:model}a). Specifically, an adapter $A_t$ is activated and trained when a new task $D_t$ emerges. Previous adapters are deactivated, allowing $A_t$ to learn in isolation and specialize in task $t$ \cite{wang-etal-2023-rehearsal}. We keep the backbone frozen and train only the adapter $A_t$ on the current task $t$.
To simplify the explanation, we omit the fact that each task uses $L$ adapters that match the number of PLM layers.

Adapters offer significant benefits \cite{wistuba2024choice}. Their robust modeling capabilities allow them to effectively capture distribution-specific details from each task and the high parameter efficiency enables efficient training of our method whenever a new task emerges. 

\paragraph{Memory}

Our method is equipped with a non-parametric memory $M = \{(x_i,y_i)\}$ that stores input-output pairs of previously seen training examples. A similar component is used in replay methods \cite{NEURIPS2019_f8d2e80c}, where memory is used to perform replay during training. 
However, our memory is leveraged prior to test inference to learn a routing function resembling the local adaptation process to improve generalization. 

We determine the memory capacity as a percentage of the total dataset size and populate it by randomly sampling examples from the training data $ M = \{\texttt{sample}(D_1), ..., \texttt{sample}(D_T) \} $. 
Random sampling effectively captures the global distribution of the data stream and has proven successful in experience replay \cite{araujo-etal-2022-relevant}, so we hypothesize that it will similarly support router training.
While more advanced memory population methods could enhance router performance \cite{Hurtado_2023_ICCV,hayes2021selective}, we leave this exploration for future work.

\paragraph{Memory-based Router Learning}

Once the adapters $A$ have been trained, they can be combined for inference (\autoref{figure:model}b). For this, we implement a router network $R$ within the model. Note that, for simplicity, we omit the detail that a separate $R$ is added for each of the $L$ layers.
This router computes the probability of the input being sent to each adapter.
This provides two advantages: 1) the routers learn to route based on input structure, and 2) can produce different combinations at different layers based on the model's abstraction hierarchy.

In practice, the router takes the [\textsc{cls}] representation as input to generate an allocation vector $z = R(h_{[\textsc{cls}]})$, which is then used to compute the output distribution. Rather than applying Softmax across adapters, where modules compete for activation, we adopt an approach inspired by \citet{ponti-etal-2023-combining}, emphasizing task-level modularity that respects task hierarchy, with more complex tasks encompassing simpler ones as sub-tasks. In this case, the router would output a binary scalar for each adapter, indicating whether it is active for a given input. As a binary vector is not differentiable, this is implemented as a collection of Bernoulli distributions, relaxed through a Gumbel-sigmoid to ensure stochasticity: $\hat{z}_{t} = \sigma \log \left[ \frac{\sigma(z_t) u}{(1 - \sigma(z_t)) (1 - u)} \right]$, where $u \sim \texttt{Uniform}(0,1)$.

To obtain a competent router, we propose a memory-based router learning process. Inspired by local adaptation, we leverage the elements in memory to train the router network parameters. However, unlike local adaptation, which adjusts the model for every inference example, our method performs this process only once after the training phase to learn the routing functions, enabling inference on any example without further adaptation.

\paragraph{Adapter Composition}
Given routing probabilities, the model can compose its adapters in different ways. We consider two options: 1) a weighted average of their outputs \cite{shazeer2017outrageously}, denoted as L2R-wavg, and 2) merging adapter parameters via a weighted average of their weights \cite{pmlr-v162-wortsman22a}, referred to as L2R-merge.

Note that these approaches differ slightly. The first uses all adapters for inference and returns a weighted average of the hidden state outputs: $H_{wavg} = \sum_{t=1}^{T} \hat{z}_t * H_t$, where $H_t = A_t(x)$, while the second merges all adapters into a single one to be used to produce a unique hidden state output: $A_{merge} = \sum_{t=1}^{T} \hat{z}_t * A_t$.

\section{Experimental Setup}
\paragraph{Benchmarks}
We adopt two CL setups: 1) Class-Incremental Learning (CIL) and 2) Task-Incremental Learning (TIL). Both setups aim to incrementally learn new tasks (and thus new classes), but TIL always has access to task identity, making CIL a more challenging and realistic scenario.

Based on this, we consider three benchmarks. \textit{MTL5} \cite{NEURIPS2019_f8d2e80c}: 5 text classification tasks. \textit{WOS} \cite{8260658}: 7 document classification tasks. \textit{AfriSenti} \cite{muhammad-etal-2023-afrisenti}: 12 (multilingual) sentiment analysis tasks. 
More details and training orders are in Appendix~\ref{sec:appendix:data}.

\paragraph{Baselines}
We focus our experimentation on comparing PEFT-based CL methods.
\textit{Lower-Bound} trains task-specific adapters, summing their outputs during inference.
\textit{Upper-Bound} trains task-specific adapters, using only the corresponding adapter for inference.
\textit{ProgPrompt} \cite{razdaibiedina2023progressive} progressively adds new prompts for new tasks. All prompts are used for inference (only for TIL).
\textit{DAM} \cite{cheng2024dynamic} learns task-specific adapters and creates task vectors by averaging training example representations. At inference, it uses similarity scores between the input and task vectors to route (only for CIL). 
\textit{EPI} \cite{wang-etal-2023-rehearsal} learns task-specific prompts and records task distributions. At inference, it selects the nearest task's prompt based on input comparison with all distributions.
\textit{MoCL} \cite{wang2024rehearsalfree} progressively adds new prompts and learns a task vector. At inference, it uses task vectors to compute similarity scores to combine the prompts.

\paragraph{Implementation Details}

We use BERT \cite{devlin-etal-2019-bert} for MTL5 and WOS while AfroXLMR \cite{alabi-etal-2022-adapting} for AfriSenti. To implement our adapters, We adopt LoRA \cite{hu2022lora} due to its efficiency and performance.
Our memory is similar to that of \cite{NEURIPS2019_f8d2e80c}, but we accumulate only 10\% of each task's, training dataset 
a negligible amount when using efficient storage formats.
Our router network consists of a linear transformation followed by a Gumbel-sigmoid \cite{maddison2017the}. 
In CIL, routers are trained using all memory examples to route inputs universally across tasks. In TIL, leveraging task identity, routers are trained with task-specific memory elements. For more details, see Appendix~\ref{sec:appendix:model}.

\begin{table}[]
\centering
\small
\begin{tabular}{clccc}
\hline
\textbf{Setup} & \multicolumn{1}{l}{\textbf{Method}} & \multicolumn{1}{c}{\textbf{MTL5}} & \multicolumn{1}{c}{\textbf{WOS}} & \multicolumn{1}{c}{\textbf{AfriSenti}} \\
\hline
& Lower-Bound                         & 16.2                              & 15.07                            & 33.64                                  \\
& Upper-Bound                         & 79.4                              & 90.06                            & 62.16                                  \\
\hline
& DAM                                 & 27.4                              & 10.00                            & 48.78                                  \\
& EPI$\dagger$                                 & 77.3                              & 77.83                            & 43.10                                  \\
\textbf{CIL} & MoCL$\dagger$                                & 73.8                              & 79.23                            & 45.61                                  \\
& L2R-wavg                           & \textbf{78.0}                     & 79.90                            & \textbf{60.04}                         \\
& L2R-merge                          & 77.7                              & \textbf{79.98}                   & 52.82                                 \\
\hline
& ProgPrompt$\dagger$                          & 77.9          & 89.93          & 49.07              \\
& EPI$\dagger$                                 & 77.3          & 77.83          & 43.10              \\
\textbf{TIL} & MoCL$\dagger$                                & \textbf{79.4} & \textbf{90.59} & 56.77              \\
& L2R-wavg                           & \textbf{79.4} & 89.24          & \textbf{62.62}     \\
& L2R-merge                          & 79.3          & 89.16          & 53.97             \\
\hline
\end{tabular}
\caption{Accuracy results on MTL5, WOS, and AfriSenti benchmarks for (top) CIL and (bottom) TIL setups. We present the average performance across all orders. $\dagger$ indicates results from \cite{wang2024rehearsalfree}.}
\label{tab:results}
\end{table}

\begin{figure}[t]
\centering
\includegraphics[width=\columnwidth]{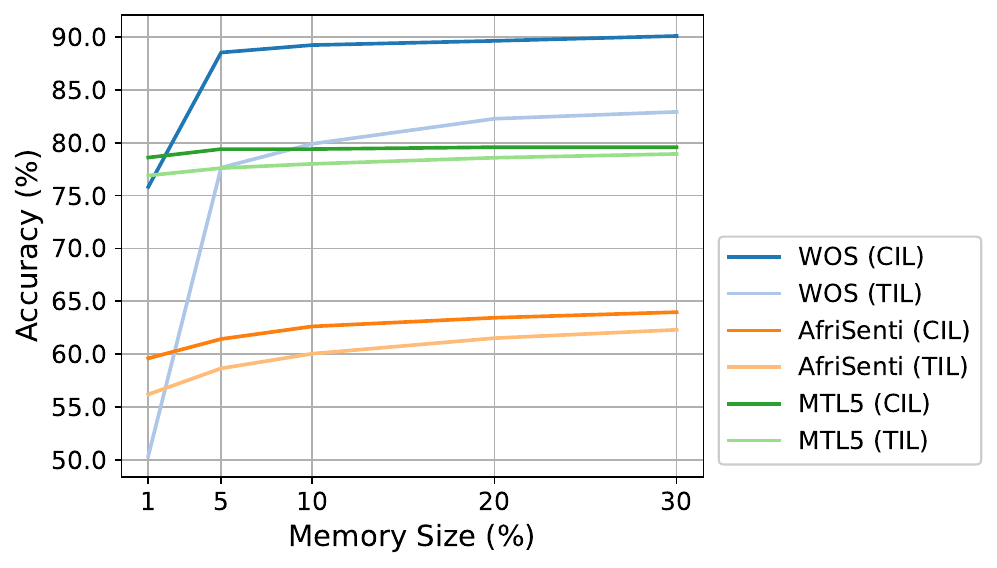}
\vspace{-0.5cm}
\caption{L2R-wavg performance across tasks and CL setups. Results for order 3 are shown for MTL5 and AfriSenti, and for order 1 for WOS.}
\label{fig:memsize}
\end{figure}

\section{Results}

In this section, we present our results, summarized in \autoref{tab:results}, which reports the average performance across orders for each benchmark. Full results are provided in Appendix~\ref{sec:appendix:results}.
Additionally, a brief efficiency analysis is available in ~\ref{sec:appendix:efficiency}. 

\paragraph{Results on CIL}
\autoref{tab:results} presents the results under the CIL setup. Both L2R-wavg and L2R-merge outperform comparable baselines such as DAM, EPI, and MoCL. DAM shows the worst performance of the group, as it relies on task vectors created from training data, which are insufficient for properly routing the adapters. However, for Afrisenti, DAM achieves competitive results since this benchmark resembles a domain incremental learning scenario where task vectors effectively help distinguish task identity. EPI and MoCL perform better than DAM but still fall short of our method despite their sophisticated techniques. This highlights the importance of a well-learned router, which enables the router’s ability to intelligently route and confer a significant performance advantage \cite{dikkala-etal-2023-benefits}.

Regarding L2R versions, we observe that L2R-wavg has a clear advantage over L2R-merge. This may be due to interference between the parameters of multiple adapters, which can cause performance drops when they are merged \cite{10.5555/3666122.3666432}.

\paragraph{Results on TIL}
Table~\ref{tab:results} presents the results under the TIL setup. L2R-wavg performs best in the MTL5 and AfriSenti benchmarks, while it lags behind in the WOS benchmark. Interestingly, both L2R-wavg and MoCL reach the Upper-Bound performance, suggesting that the learned routing function likely emulates the full activation of the corresponding adapter or an effective combination of the adapters. We hypothesize that L2R-wavg tends toward the latter, as evidenced by its performance in AfriSenti, where it surpasses the Upper-Bound.

In the WOS benchmark, our methods perform competitively but fall behind MoCL. We attribute this to the amount of data used to train the routing function. WOS consists of very small datasets, resulting in a memory populated with approximately 100 elements per task, which may not be sufficient for router learning. In fact, \citet{dikkala-etal-2023-benefits} have shown that as the number of examples increases, the routing function reaches its optimal performance. We explore this in the next section.

\begin{table}[]
\centering
\small
\begin{tabular}{lccc}
\hline
\multicolumn{1}{l}{\textbf{Method: L2R-wavg}} & \multicolumn{1}{c}{\textbf{MTL5}} & \multicolumn{1}{c}{\textbf{WOS}} & \multicolumn{1}{c}{\textbf{AfriSenti}} \\
\hline
Gumbel-sigmoid                          & \textbf{78.0}                     & \textbf{79.90}                            & \textbf{60.04}                         \\
Softmax                          & 77.8                              & 60.74                   & 59.90                                 \\
\hline
\end{tabular}
\caption{Accuracy results on MTL5, WOS, and AfriSenti benchmarks for the CIL setup of L2R-wavg using Gumbel-sigmoid and Softmax.}
\label{tab:gumbelvssoftmax}
\vspace{-0.25cm}
\end{table}

\begin{figure*}[t]
\centering
\includegraphics[width=\linewidth]{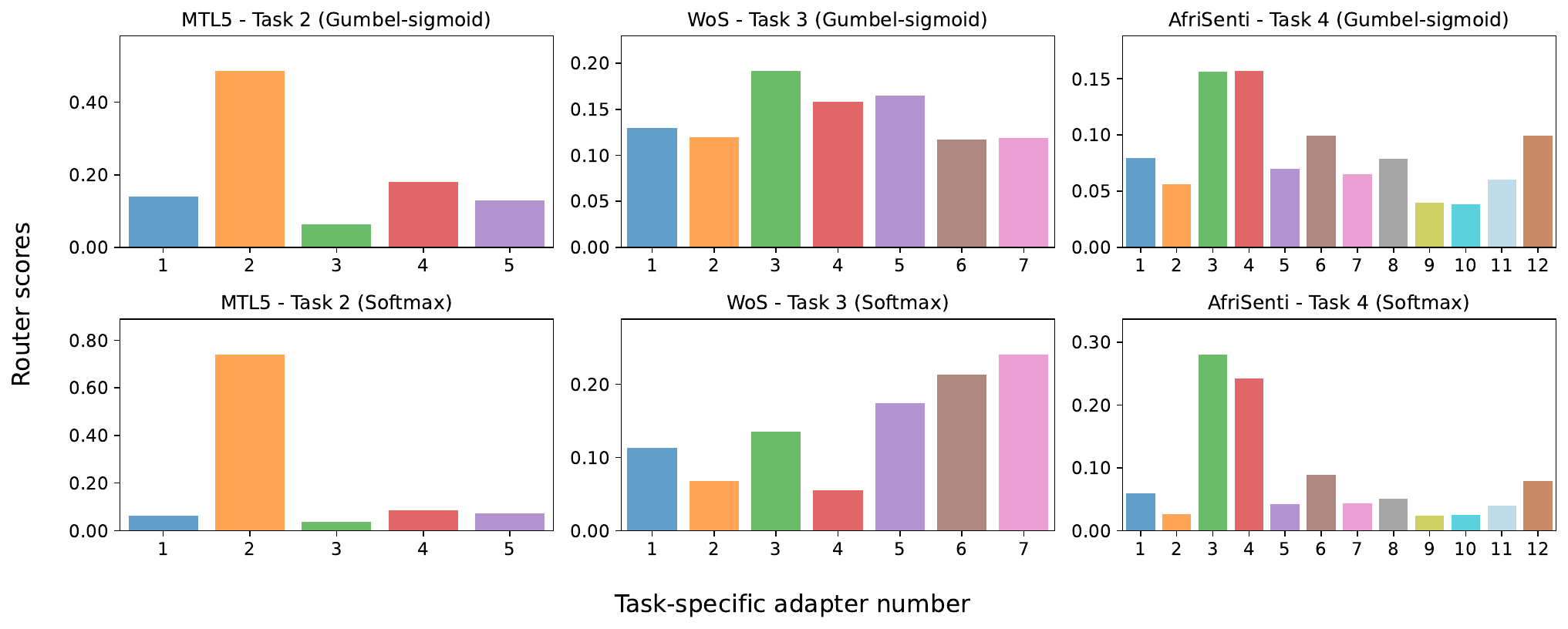}
\caption{Average router scores for task 2 of MTL5 (order 4), task 3 of WOS (order 1), and task 4 of AfriSenti (order 1) using Gumbel-sigmoid (top) and Softmax (bottom). Scores were computed on the test sets.}
\label{figure:routerscore}
\end{figure*}

\paragraph{Impact of Memory Size} 

\autoref{fig:memsize} shows the performance for all benchmarks with memory sizes ranging from 1\% to 30\%. As expected, we find that the performance of L2R-wavg increases with larger memory sizes. Notably, the performance on WOS improves to match that of MoCL, supporting our assertion about the importance of using more examples for effective router adaptation.
For MTL5 and AfriSenti, their performance gets close to the Upper-Bound, demonstrating that L2R may be learning better compositions for the adapters, leading to high performance across tasks.

\paragraph{Impact of Gumbel-sigmoid Distribution} 
As discussed in \autoref{sec:method}, we use Gumbel-sigmoid distribution during the training of the routers to theoretically enhance task-level modularity over activation competency offered by Softmax. To validate this, we compare the resulting routing scores from L2R-wavg in CIL with the ones from a version trained using Softmax function.

\autoref{tab:gumbelvssoftmax} shows the results, highlighting that using Gumbel-sigmoid during the training of the routing function consistently outperforms the use of Softmax activation. The improvement is observed across all benchmarks, with accuracy gains ranging from 0.1 to 19.2 points. The WOS benchmark exhibits the most significant improvement, likely because it consists of more tasks, which allows for more modules to be leveraged.

\autoref{figure:routerscore} shows router probabilities for the test sets of MTL5 (order 4), WoS (order 1), and AfriSenti (order 1). On the left, for MTL5 task 2 (news classification), adapter 2 is the most activated in both model versions. However, the Softmax version predominantly uses adapter 2, while the Gumbel-sigmoid version also leverages adapters 1 and 4, which specialize in sentiment analysis.

In the middle, for the WOS task 3, both models activate multiple adapters because all tasks are document categorization across various domains. With Gumbel-sigmoid, adapter 3 (the correct one) is primarily activated, while the Softmax version strongly activates adapter 7, with less use of others.

On the right, for AfriSenti task 4 (sentiment analysis with varying input languages), the Gumbel-sigmoid version strongly activates adapters 3 and 4, with lighter activation of the rest. Adapters 3 and 4 correspond to Hausa and Igbo, both spoken in Nigeria, explaining their activation. Adapter 4 is slightly more activated, showing the router’s ability to detect the target task while using other adapters. In contrast, the Softmax version primarily activates adapter 3, with less activation of others.

\section{Conclusion}
We introduced L2R, a method for routing in CL that enables PLMs to dynamically combine adapters, improving generalization and performance. Experiments across benchmarks demonstrate L2R's effectiveness in both TIL and the challenging CIL setting. We also show that larger memory enhances routing function learning and that L2R effectively leverages diverse adapter combinations.

\section*{Limitations}
Our method focuses on developing an effective routing function to enhance generalization and performance on downstream tasks learned from a stream. To achieve this, we use a non-parametric memory to store previous examples, similar to replay-based methods. This approach may be limited in environments with storage constraints or data privacy concerns.

Our experimental setup primarily focuses on small-scale PLMs, which is a limitation since exploring large language models (LLMs) would be desirable given their dominance in the current NLP landscape. Although we have not tested our method with LLMs due to computational resource limitations, we anticipate similar results.

\section{Acknowledgements}

This work was funded by the European Research Council (ERC) under the European Union’s Horizon 2020 research and innovation programme: CALCULUS project (Grant Agreement No. 788506) and KeepOnLearning project (Grant Agreement No. 101021347).

\bibliography{custom}

\clearpage

\appendix

\section{Additional Benchmark Details}
\label{sec:appendix:data}

MTL5 is a benchmark for text classification. It consists of five datasets from \citep{NIPS2015_250cf8b5}. AGNews classification, Yelp sentiment analysis, Amazon sentiment analysis, DBPedia article classification, and Yahoo questions and answers categorization. Both sentiment analysis tasks share the same labels. In line with \citet{NEURIPS2019_f8d2e80c}, we use 575,000 training and 38,000 test examples with 33 classes from all datasets using 4 task orders:

\begin{enumerate}[(i)]
    \small
    \itemsep-0.25em
    \item AGNews → Yelp → Amazon → Yahoo → DBpedia
    \item Yelp → Yahoo → Amazon → DBpedia → AGNews
    \item DBPedia → Yahoo → AGNews → Amazon → Yelp
    \item Yelp → AGNews → DBPedia → Amazon → Yahoo
\end{enumerate}

Web-of-science (WOS) \cite{8260658} initially began as a hierarchical dataset for categorizing documents. It includes research papers from seven distinct fields: biochemistry, civil engineering, computer science, electrical engineering, medical science, mechanical engineering, and psychology. Each of these fields represents a high-level category for document classification, with multiple subcategories within each field. 
This dataset has about 11,967 instances and 33 classes. In line with \citet{wang-etal-2023-rehearsal}, we split to train/val/test set in the ratio of 0.6:0.2:0.2 and structured 7 sequential learning tasks based on the high-level categories of the dataset:

\begin{enumerate}[(i)]
    \small
    \itemsep-0.25em 
    \item 1 → 2 → 3 → 4 → 5 → 6 → 7
\end{enumerate}

AfriSenti \cite{muhammad-etal-2023-afrisenti} is a multilingual sentiment analysis dataset comprising 12 low-resource African languages. These languages include Amharic (am), Algerian Arabic (dz), Hausa (ha), Igbo (ig), Kinyarwanda (kr), Moroccan Arabic (ma), Nigerian Pidgin (pcm), Mozambican Portuguese (pt), Swahili (sw), Xitsonga (ts), Twi (twi), and Yoruba (yo). The dataset contains over 110,000 annotated tweets and spans three sentiment classes across all languages.
In line with \cite{wang-etal-2023-rehearsal}, we use 3 task orders:

\begin{enumerate}[(i)]
    \small
    \itemsep-0.25em 
    \item am → dz → ha → ig → kr → ma → pcm → pt → sw → ts → twi → yo
    \item ma → pcm → kr → pt → ig → sw → ha → ts → dz → twi → am → yo
    \item am → dz → ha → ma → ig → kr → sw → ts → twi → yo → pcm → pt
\end{enumerate}

Following \citet{wang-etal-2023-rehearsal}, we categorize benchmarks based on task domain similarity. WOS and Afrisenti serve as near-domain benchmarks due to their closely related tasks. MTL5, on the other hand, represents far-domain benchmarks where distinct domain boundaries among tasks are evident.

\section{Additional Implementation Details}
\label{sec:appendix:model}

As backbones, we use BERT-base \cite{devlin-etal-2019-bert} for MTL5 and WOS and AfroXLMR-base \cite{alabi-etal-2022-adapting} for AfriSenti. Additionally, we adopt LoRA \cite{hu2022lora} as our PEFT modules, using a rank of 8 and dropout of 0.1. We applied LoRA to $W_q$ and $W_v$, and it has been demonstrated to be a competitive configuration. These adapters allow efficient fine-tuning, representing less than 0.7\% of the PLM’s parameters in our experiments.

Our memory is non-parametric, which means that it stores raw examples with labels sampled from training examples. We only sample 10\% of each task's training dataset, representing less than 3.8MB per dataset in our experiments, as we use Parquet, an efficient storage format.

During the training phase, we use a batch size of 8 and a learning rate of 3e-4. Additionally, we use a linear scheduler with warmup and AdamW optimizer.
On the other hand, we use the same configuration to train the router networks. However, we use a learning rate of 3e-4 or 3e-5 for MTL5 and AfriSenti, and a learning rate of 3e-3 or 3e-4 for WOS.

\section{Full Results}
\label{sec:appendix:results}
\autoref{tab:cil} and \autoref{tab:til} show the full results for CIL and TIL setups, respectively.
We observe that our method consistently outperforms the best baseline (MoCL) across orders in CIL. Interestingly, our model achieves a similar performance across orders. This is because the isolation strategy helps the model to have a very specialized adapter regardless of the order of training.
However, order (iii,iv) of MTL5 and order (ii) of AfriSenti show slightly superior performance, indicating that memory population methods could be leveraged to further improve performance \cite{araujo-etal-2022-relevant}.

Regarding TIL experiments, we find that L2R-wavg obtains a similar performance to MoCL and Upper-Bound. Interestingly, L2R-merge lags behind the performance of MoCL, potentially due to the need for more data to properly train its router, as shown in our experiments. This is more evident for our L2R-merge, which has also been shown to be less effective to L2R-wavg possible for interference produced by merging the adapters.

\begin{table*}[]
\centering
\begin{tabular}{lccccc}
\hline
                                    & \multicolumn{5}{c}{\textbf{MTL5}}                                                                                                    \\
\multicolumn{1}{c}{\textbf{Method}} & \textbf{i}               & \textbf{ii}               & \textbf{iii}               & \textbf{iv}               & \textbf{Avg}             \\
\hline
Lower-Bound                         & 16.2                     & 16.2                     & 16.2                     & 16.2                     & 16.2                     \\
Upper-Bound                         & 79.4                     & 79.4                     & 79.4                     & 79.4                     & 79.4                     \\
\hline
DAM                                 & 27.4                     & 27.4                     & 27.4                     & 27.4                     & 27.4                     \\
EPI$\dagger$                                 & \multicolumn{1}{l}{77.4} & \multicolumn{1}{l}{77.3} & \multicolumn{1}{l}{77.2} & \multicolumn{1}{l}{77.4} & \multicolumn{1}{l}{77.3} \\
MoCL$\dagger$                                & 73.0                     & 74.0                     & 74.8                     & 73.6                     & 73.8                     \\
L2R-wavg                            & 77.8                     & 77.9                     & 78.2                     & 78.1                     & \textbf{78.0}                     \\
L2R-merge                          & 77.7                     & 77.7                     & 77.6                     & 77.8                     & \textbf{77.7}                    \\
\hline
\end{tabular}
\begin{tabular}{c}
\hline
\textbf{WOS}              \\
\textbf{i}              \\
\hline
15.07                     \\
90.06                     \\
\hline
10.00                     \\
77.83                    \\
79.23                     \\
\textbf{79.90}                     \\
\textbf{79.98}                    \\
\hline
\end{tabular}
\begin{tabular}{cccc}
\hline
\multicolumn{4}{c}{\textbf{AfriSenti}}                                                                       \\
\textbf{i}               & \textbf{ii}                & \textbf{iii}                & \textbf{Avg}              \\
\hline
33.64                    & 33.64                     & 33.64                     & 33.64                     \\
62.16                    & 62.16                     & 62.16                     & 62.16                     \\
\hline
48.78                    & 48.78                     & 48.78                     & 48.78                     \\
42.96                    & 42.97                     & 43.36                     & 43.10                     \\
45.57                    & 44.32                     & 46.95                     & 45.61                     \\
59.14                    & 60.8                      & 60.18                     & \textbf{60.04}                     \\
53.36                    & 52.07                     & 53.03                     & \textbf{52.82}                    \\
\hline
\end{tabular}
\caption{Results on MTL5, WOS, and AfriSenti for CIL. $\dagger$ indicates that the results come from \cite{wang2024rehearsalfree}.}
\label{tab:cil}
\end{table*}

\begin{table*}[]
\centering
\begin{tabular}{lccccc}
\hline
\textbf{}                           & \multicolumn{5}{c}{\textbf{MTL5}}                                                                                                                                    \\
\multicolumn{1}{c}{\textbf{Method}} & \textbf{i}               & \textbf{ii}               & \textbf{iii}               & \textbf{iv}               & \textbf{Avg}             \\
\hline
Lower-Bound                         & 16.2                           & 16.2                           & 16.2                           & 16.2                           & 16.2                             \\
Upper-Bound                         & 79.4                           & 79.4                           & 79.4                           & 79.4                           & 79.4                             \\
\hline
ProgPrompt$\dagger$                          & 78.0                           & 77.9                           & 77.9                           & 77.9                           & 77.9                             \\
EPI$\dagger$                                 & 77.4                           & 77.3                           & 77.2                           & 77.4                           & 77.3                             \\
MoCL$\dagger$                                & 79.3                           & 79.6                           & 79.2                           & 79.4                           & \textbf{79.4}                             \\
L2R-wavg                            & 79.4                           & 79.4                           & 79.5                           & 79.4                           & \textbf{79.4}                             \\
L2R-merge                          & 79.2                           & 79.3                           & 79.4                           & 79.3                           & 79.3                            \\
\hline
\end{tabular}
\begin{tabular}{c}
\hline
\multicolumn{1}{c}{\textbf{WOS}} \\
\multicolumn{1}{c}{\textbf{i}} \\
\hline
15.07                            \\
90.06                            \\
\hline
89.93                            \\
77.83                            \\
\textbf{90.59}                            \\
89.24                            \\
89.16                           \\
\hline
\end{tabular}
\begin{tabular}{cccc}
\hline
\multicolumn{4}{c}{\textbf{AfriSenti}}                                                                                              \\
\multicolumn{1}{c}{\textbf{i}} & \multicolumn{1}{c}{\textbf{ii}} & \multicolumn{1}{c}{\textbf{iii}} & \multicolumn{1}{c}{\textbf{Avg}} \\
\hline
33.64                          & 33.64                          & 33.64                          & 33.64                            \\
62.16                          & 62.16                          & 62.16                          & 62.16                            \\
\hline
50.16                          & 46.74                          & 50.30                          & 49.07                            \\
41.49                          & 42.65                          & 45.16                          & 43.10                            \\
57.05                          & 56.52                          & 56.74                          & 56.77                            \\
62.68                          & 62.78                          & 62.39                          & \textbf{62.62}                            \\
54.06                          & 53.3                           & 54.55                          & 53.97                          \\
\hline
\end{tabular}
\caption{Results on MTL5, WOS, and AfriSenti for TIL. $\dagger$ indicates that the results come from \cite{wang2024rehearsalfree}.}
\label{tab:til}
\end{table*}

\section{Efficiency Comparison: FLOPs Analysis}
\label{sec:appendix:efficiency}

We compute the theoretical amount of FLOPs (floating-point operations) of our methods and some baselines for the MTL5 benchmark (i.e., 5 tasks). In \autoref{tab:flops}, we observe BERT for classification as the more efficient model. This is expected as all the other methods augment BERT with PEFT modules. Our methods are the second most efficient models. L2R-wave is the less efficient one, as it always uses all the adapters and composes the input them. L2R-merge and DAM have the same amount of FLOPs as these models merge the adapters in one, making the computation more efficient. Although MoCL is the direct competitor to our methods in performance, it is not in terms of efficiency as this model extends the input, resulting in more operations.

\begin{table}[hbt!]
    \centering
    \begin{tabular}{ccc}
    \hline
        \textbf{Method} & \textbf{FLOPs} \\
    \hline
        BERT & 325.42 GFLOPS \\
        DAM & 326.49 GFLOPS \\
        MoCL & 477.85 GFLOPS \\
        L2R-wavg & 330.80 GFLOPS \\
        L2R-merge & 326.49 GFLOPS \\
    \hline
    \end{tabular}
    \caption{Theoretical FLOPs when processing a batch size 1 and a sequence of 128 tokens.}
    \label{tab:flops}
\end{table}

\end{document}